\documentclass{article}
\usepackage{graphicx}
\usepackage{physics}
\usepackage{amsmath}
\usepackage{amssymb}
\usepackage{mathtools}
\usepackage{a4wide}
\usepackage{authblk}
\usepackage{xurl}

\usepackage{tikz} 
\usepackage{pgfplots}
\usepackage{hyperref}
\pgfplotsset{compat=1.18}

\usetikzlibrary{shapes.geometric,decorations,backgrounds}
\usetikzlibrary{arrows.meta}
\usetikzlibrary{chains,arrows}
\usetikzlibrary{automata, positioning, arrows}
\usepackage[outline]{contour} 
\usetikzlibrary{calc}
\usetikzlibrary{angles,quotes} 
\contourlength{1.2pt}
\usetikzlibrary{decorations.pathmorphing}
\usetikzlibrary{fit}

\tikzset{snake it/.style={decorate, decoration=snake}}

\tikzset{
        >=latex, 
        node distance=2cm, 
        }
        
\definecolor{myred}{RGB}{169,44,31}
\definecolor{mygreen}{RGB}{100,200,55}
\definecolor{myblue}{RGB}{58,132,186}
\definecolor{myyellow}{RGB}{243,192,105}
\definecolor{mygray}{RGB}{242,242,242}

\title{Ornstein-Uhlenbeck Adaptation as a Mechanism for Learning in Brains and Machines} 

\date{}

\author[]{Jesús García Fernández}
\author[]{Nasir Ahmad}
\author[]{Marcel van Gerven}

\affil[1]{Department of Machine Learning and Neural Computing, Donders Institute for Brain, Cognition and Behaviour\\
  Radboud University, Nijmegen, the Netherlands}

\begin{document}
\maketitle

\begin{abstract}
Learning is a fundamental property of intelligent systems, observed across biological organisms and engineered systems. While modern intelligent systems typically rely on gradient descent for learning, the need for exact gradients and complex information flow makes its implementation in biological and neuromorphic systems challenging. This has motivated the exploration of alternative learning mechanisms that can operate locally and do not rely on exact gradients. In this work, we introduce a novel approach that leverages noise in the parameters of the system and global reinforcement signals. Using an Ornstein-Uhlenbeck process with adaptive dynamics, our method balances exploration and exploitation during learning, driven by deviations from error predictions, akin to reward prediction error. Operating in continuous time, Ornstein-Uhlenbeck adaptation (OUA) is proposed as a general mechanism for learning in dynamic, time-evolving environments. We validate our approach across a range of different tasks, including supervised learning and reinforcement learning in feedforward and recurrent systems. Additionally, we demonstrate that it can perform meta-learning, adjusting hyper-parameters autonomously. Our results indicate that OUA provides a viable alternative to traditional gradient-based methods, with potential applications in neuromorphic computing. It also hints at a possible mechanism for noise-driven learning in the brain, where stochastic neurotransmitter release may guide synaptic adjustments.
\end{abstract}



\section{Introduction}

One of the main properties of any intelligent system is that it has the capacity to learn. This holds for biological systems, ranging from bacteria and fungi to plants and animals~\cite{fernando2009molecular, gagliano2016learning, money2021hyphal, sasakura2013behavioral}, as well as for engineered systems designed by artificial intelligence (AI) researchers~\cite{sutton2018reinforcement, LeCun2015, brunton2022data}. Modern intelligent systems, such as those used in machine learning, typically rely on gradient descent for learning by minimizing error gradients~\cite{Linnainmaa1970, Werbos1974, rumelhart1985learning}. While gradient-based methods have driven significant advances in AI~\cite{LeCun2015}, their reliance on exact gradients, centralized updates, and complex information pathways limits their applicability in biological and neuromorphic systems

In contrast, biological learning likely relies on different mechanisms, as organisms often lack the exact gradient information and centralized control that gradient descent requires~\cite{lillicrap2020backpropagation, whittington2019theories}. Neuromorphic computing, inspired by these principles, aims to replicate the distributed, energy-efficient learning of biological systems~\cite{Mead1990, Modha2011}. However, integrating traditional gradient-based methods into neuromorphic hardware has proven challenging, highlighting a critical gap: the need for gradient-free learning mechanisms that exclusively rely on operations that are local in space and time~\cite{Jaeger2023,davies2021advancing}.

To address this, alternative learning principles to gradient descent have been proposed for both rate-based~\cite{oja1982simplified, bienenstock1982theory, scellier2017equilibrium, bengio2014auto, whittington2017approximation} and spike-based models~\cite{markram1997regulation, bellec2020solution, haider2021latent, payeur2021burst}. A class of methods that leverages inherent noise present in biological systems to facilitate learning is perturbation-based methods~\cite{spall1992multivariate, widrow199030, werfel2003learning}, which adjust the system's parameters based on noise effects and global reinforcement signals, offering gradient-free, local learning suitable for biological or neuromorphic systems. Specifically, these methods inject random fluctuations (noise) into the system and evaluate the impact of this noise on performance through a reinforcement signal. Then, they adjust the system's parameters to increase or decrease alignment with the output of the system featuring noise, depending on the feedback provided by the reinforcement signal. Within this framework, node perturbation methods~\cite{flower1992summed, hiratani2022stability, williams1992simple, werfel2003learning, fiete2006gradient, dalm2024efficient, Fernandez2024} injects noise into the nodes, while weight perturbation methods~\cite{kirk1992analog, lippe1994study, zuge2023weight, cauwenberghs1992fast, dembo1990model} inject noise into the parameters. Particularly, node perturbation has been shown to approximate gradient descent updates on average~\cite{werfel2003learning}, but it relies on a more stochastic exploration of the loss landscape rather than consistently following the steepest descent direction, as in gradient-based methods. Nevertheless, both approaches require generating two outputs per input -- one noisy and one noise-free -- to accurately measure the impact of the noise on the system, and access to the noise process, making them impractical in many real-world or biological scenarios. Reward-modulated Hebbian learning (RMHL)~\cite{legenstein2010reward, miconi2017biologically}, a bio-inspired alternative overcomes some of these limitations as it does not need a noiseless system nor access to the noise process to learn. Yet, RMHL often struggles to solve practical tasks efficiently, limiting their applicability.

In this work, we propose Ornstein-Uhlenbeck Adaptation (OUA), a novel learning mechanism that extends the strengths of perturbation-based methods while addressing some of their limitations. Like RMHL methods, it does not require a noiseless system or direct access to the noise process to learn. Differently, OUA introduces a mean-reverting Ornstein-Uhlenbeck (OU) process~\cite{uhlenbeck1930theory, Doob1942Brownian} to inject noise directly into system parameters, adapting their mean based on a global modulatory reinforcement signal. This signal is derived from deviations in error predictions, resembling a reward prediction error (RPE) thought to play a role in biological learning~\cite{schultz1997neural}. Unlike traditional methods, OUA operates in continuous time using differential equations, making it particularly suited for dynamic, time-evolving environments~\cite{Kudithipudi2022}. While RMHL was developed to model biological learning dynamics, OUA’s online nature and ability to adapt continually offer advantages in practical applications. Additionally, the simple and local nature of the proposed mechanism makes it well-suited for implementation on neuromorphic hardware.

We validate our approach across various experiments, including feedforward and recurrent systems, covering input-output mappings, control tasks, and a real-world weather forecasting task, all within a continuous-time framework. Additionally, we demonstrate that the method can be extended to a meta-learning setting by learning the system hyper-parameters. Finally, we discuss the implications of OUA, as our experiments demonstrate its efficiency and versatility, positioning it as a promising approach for both neuromorphic computing and understanding biological learning mechanisms.

\section{Methods}

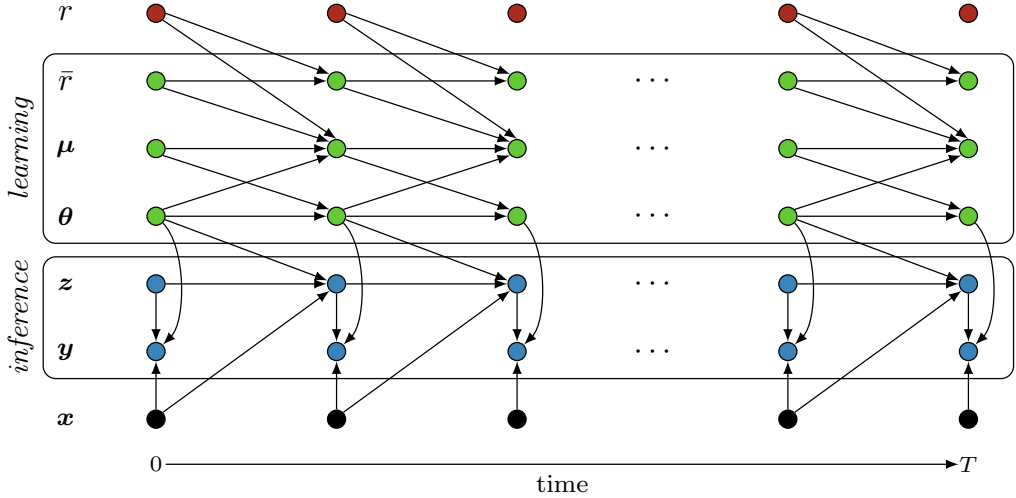
\begin{figure}[!ht]
\centering
\scalebox{1.2}{
\begin{tikzpicture}[>=Stealth, node distance=2cm, minimum size=0.2cm, inner sep=0pt]
        
    \node (x) at (-1,-0.5) {$\footnotesize{\vb*{x}}$};
    \node (x0) [draw=black, circle, fill=black] at (0,-0.5) {};
    \node (x1) [draw=black, circle, fill=black] at (2,-0.5) {};
    \node (x2) [draw=black, circle, fill=black] at (4,-0.5) {};
    \node (xn) [draw=black, circle, fill=black] at (7,-0.5) {};
    \node (xN) [draw=black, circle, fill=black] at (9,-0.5) {};

    \node (t0) at (0,-1.0) {\scriptsize{$0$}};
    \node (tm) [label=below:\footnotesize{time}] at (4.5,-1.0) {};
    \node (tN) at (9,-1.0) {\scriptsize{$T$}};
    \draw[->, >=latex] (t0) -- (tN);

    \node (z) at (-1,1) {$\footnotesize{\vb*{z}}$};
    \node (z0) [draw=black, circle, fill=myblue] at (0,1) {};
    \node (z1) [draw=black, circle, fill=myblue] at (2,1) {};
    \node (z2) [draw=black, circle, fill=myblue] at (4,1) {};
    \node (zdots) at (5.5,1){$\quad\cdots\quad$};
    \node (zn) [draw=black, circle, fill=myblue] at (7,1){};
    \node (zN) [draw=black, circle, fill=myblue] at (9,1) {};


    \node (p) at (-1,1.75) {$\footnotesize{\vb*{\theta}}$};
    \node (p0) [draw=black, circle, fill=mygreen] at (0,1.75) {};
    \node (p1) [draw=black, circle, fill=mygreen] at (2,1.75) {};
    \node (p2) [draw=black, circle, fill=mygreen] at (4,1.75) {};
    \node (pdots) at (5.5,1.75){$\quad\cdots\quad$};
    \node (pn) [draw=black, circle, fill=mygreen] at (7,1.75){};
    \node (pN) [draw=black, circle, fill=mygreen] at (9,1.75) {};

    \node (m) at (-1,2.5) {$\footnotesize{\vb*{\mu}}$};
     \node (m0) [draw=black, circle, fill=mygreen] at (0,2.5) {};
    \node (m1) [draw=black, circle, fill=mygreen] at (2,2.5) {};
    \node (m2) [draw=black, circle, fill=mygreen] at (4,2.5) {};
    \node (mdots) at (5.5,2.5){$\quad\cdots\quad$};
    \node (mn) [draw=black, circle, fill=mygreen] at (7,2.5){};
    \node (mN) [draw=black, circle, fill=mygreen] at (9,2.5) {};

    \node (rb) at (-1,3.25) {$\footnotesize{\bar{r}}$};
    \node (rb0) [draw=black, circle, fill=mygreen] at (0,3.25) {};
    \node (rb1) [draw=black, circle, fill=mygreen] at (2,3.25) {};
    \node (rb2) [draw=black, circle, fill=mygreen] at (4,3.25) {};
    \node (rbdots) at (5.5,3.25){$\quad\cdots\quad$};
    \node (rbn) [draw=black, circle, fill=mygreen] at (7,3.25){};
    \node (rbN) [draw=black, circle, fill=mygreen] at (9,3.25) {};

    \node (y) at (-1,0.25) {$\footnotesize{\vb*{y}}$};
    \node (y0) [draw=black, circle, fill=myblue] at (0,0.25) {};
    \node (y1) [draw=black, circle, fill=myblue] at (2,0.25) {};
    \node (y2) [draw=black, circle, fill=myblue] at (4,0.25) {};
    \node (ydots) at (5.5,0.25){$\quad\cdots\quad$};

    \node (yn) [draw=black, circle, fill=myblue] at (7,0.25) {};
    \node (yN) [draw=black, circle, fill=myblue] at (9,0.25) {};

    \node (r) at (-1,4) {$\footnotesize{r}$};
    \node (r0) [draw=black, circle, fill=myred] at (0,4) {};
    \node (r1) [draw=black, circle, fill=myred] at (2,4) {};
    \node (r2) [draw=black, circle, fill=myred] at (4,4) {};
    \node (rn) [draw=black, circle, fill=myred] at (7,4) {};
    \node (rN) [draw=black, circle, fill=myred] at (9,4) {};

    \node [rotate=90] (L) at (-1.5,2.5) 
    {\small{\it learning}};

    \node [rotate=90] (I) at (-1.5,0.25 + 0.375) {\small{\it inference}};

        \scoped[on background layer]
    \draw[rounded corners, draw] (p0)+(-1.25,-0.3) rectangle ([xshift=0.5cm,yshift=0.3cm]rbN) {};

     \scoped[on background layer]
    \draw[rounded corners, draw] (y0)+(-1.25,-0.3) rectangle ([xshift=0.5cm,yshift=0.3cm]zN) {};

 \draw[->, >=latex] (z0) -- (z1);
    \draw[->, >=latex] (z1) -- (z2);
    \draw[->, >=latex] (zn) -- (zN);
      
   \draw[->, >=latex] (m0) -- (m1);
    \draw[->, >=latex] (m1) -- (m2);
    \draw[->, >=latex] (mn) -- (mN);
    
     \draw[->, >=latex] (p0) -- (p1);
    \draw[->, >=latex] (p1) -- (p2);
    \draw[->, >=latex] (pn) -- (pN);

 \draw[->, >=latex] (rb0) to [out=-45,in=135,looseness=0] (m1);
 \draw[->, >=latex] (rb1) to [out=-45,in=135,looseness=0] (m2);
\draw[->, >=latex] (rbn) to [out=-45,in=135,looseness=0] (mN);

 \draw[->, >=latex] (rb0) -- (rb1);
    \draw[->, >=latex] (rb1) -- (rb2);
    \draw[->, >=latex] (rbn) -- (rbN);

      \draw[->, >=latex] (m0) to [out=-45,in=135,looseness=0] (p1);
    \draw[->, >=latex] (m1) to [out=-45,in=135,looseness=0] (p2);
    \draw[->, >=latex] (mn) to [out=-45,in=135,looseness=0] (pN);

  \draw[->, >=latex] (p0) to [out=45,in=-135,looseness=0] (m1);
    \draw[->, >=latex] (p1) to [out=45,in=-135,looseness=0] (m2);
    \draw[->, >=latex] (pn) to [out=45,in=-135,looseness=0] (mN);
    
    \draw[->, >=latex] (x0) to [out=45,in=-135,looseness=0] (z1);
    \draw[->, >=latex] (x1) to [out=45,in=-135,looseness=0] (z2);
    \draw[->, >=latex] (xn) to [out=45,in=-135,looseness=0] (zN);
    
       \draw[->, >=latex] (z0) -- (y0);
       \draw[->, >=latex] (z1) -- (y1);
       \draw[->, >=latex] (z2) -- (y2);
       \draw[->, >=latex] (zn) -- (yn);
       \draw[->, >=latex] (zN) -- (yN);
   
       \draw[->, >=latex] (x0) to [out=90,in=-90,looseness=0.7] (y0);
       \draw[->, >=latex] (x1) to [out=90,in=-90,looseness=0.7] (y1);
       \draw[->, >=latex] (x2) to [out=90,in=-90,looseness=0.7] (y2);
       \draw[->, >=latex] (xn) to [out=90,in=-90,looseness=0.7] (yn);
       \draw[->, >=latex] (xN) to [out=90,in=-90,looseness=0.7] (yN);

       \draw[->, >=latex] (p0) to [out=-45,in=45,looseness=0.7] (y0);
       \draw[->, >=latex] (p1) to [out=-45,in=45,looseness=0.7] (y1);
       \draw[->, >=latex] (p2) to [out=-45,in=45,looseness=0.7] (y2);
       \draw[->, >=latex] (pn) to [out=-45,in=45,looseness=0.7] (yn);
       \draw[->, >=latex] (pN) to [out=-45,in=45,looseness=0.7] (yN);

    \draw[->, >=latex] (p0) -- (z1);
    \draw[->, >=latex] (p1) -- (z2);
    \draw[->, >=latex] (pn) -- (zN);

    \draw[->, >=latex] (r0) to [out=0,in=135,looseness=0] (rb1);
    \draw[->, >=latex] (r1) to [out=0,in=135,looseness=0] (rb2);
    \draw[->, >=latex] (rn) to [out=0,in=135,looseness=0] (rbN);

    \draw[->, >=latex] (r0) to [out=-45,in=90,looseness=0] (m1);
    \draw[->, >=latex] (r1) to [out=-45,in=90,looseness=0] (m2);
    \draw[->, >=latex] (rn) to [out=-45,in=90,looseness=0] (mN);

\end{tikzpicture}
}\caption{Dependency structure of the variables that together determine Ornstein-Uhlenbeck adaptation (hyper-parameters not shown). Variables $\bar{r}$, $\vb*{\mu}$ and $\vb*{\theta}$ (green) are related to learning whereas variables $\vb*{z}$ and $\vb*{y}$ (blue) are related to inference.  
The average reward estimate $\bar{r}$ depends on rewards $r$ (red) that indirectly depend on the outputs $\vb*{y}$ generated by the model. The output itself depends on input $\vb*{x}$ (black).}
\label{fig:SSM}
\end{figure}

\subsection{Inference}

Consider an inference problem, where the goal is to map inputs $\vb*{x}(t) \in \mathbb{R}^m$ to outputs $\vb*{y}(t) \in \mathbb{R}^k$ for $t \in \mathbb{R}^+$.  In a continual learning setting, this problem can be formulated as a stochastic state space model
\begin{align}
\dd\vb*{z}(t) 
&= f_{\vb*{\theta}}(\vb*{z}(t), \vb*{x}(t)) \dd{t} + \dd{\vb*{\zeta}}(t)\label{eq:sssm1}
\\
\vb*{y}(t) &= g_{\vb*{\theta}}(\vb*{z}(t), \vb*{x}(t)) + \vb*{\epsilon}(t)
\label{eq:sssm2}
\end{align}
Here, $\vb*{\theta} \in \mathbb{R}^n$ are (learnable) parameters, $\vb*{z}(t) \in \mathbb{R}^d$ is a latent process, $f_{\vb*{\theta}}(\cdot)$ and $g_{\vb*{\theta}}(\cdot)$ are nonlinear functions parameterized by $\vb*{\theta}$, $\vb*{\zeta}(t)$ is process noise, and $\vb*{\epsilon}(t)$ is observation noise. We assume that we integrate the process from some initial time $t_0$ up to some time horizon $T$. We also refer to Equation~\eqref{eq:sssm1} as a (latent) neural stochastic differential equation~\cite{tzen2019neural}, where `neural' refers to the use of learnable parameters $\vb*{\theta}$.

\subsection{Reward prediction}

To enable learning, we assume the existence of a global scalar reward signal $r(t)$, providing instantaneous feedback on the efficacy of the system's output $y(t)$. The goal is to adapt the parameters $\vb*{\theta}$ to maximize the cumulative reward or return
\begin{equation}
G(t) = \int_{t_0}^t r(\tau) \dd{\tau}
\end{equation}
as $t \rightarrow T$. This can be expressed as an update equation $\dd{G}(t) = r(t) \dd{t}$ with initial state $G_0 = G(t_0) = 0$. To facilitate learning, the system maintains a moving average of the reward~\cite{wan2021learning} according to
\begin{equation}
\dd{\bar{r}}(t) = \rho (r(t) - \bar{r}(t)) \dd{t}.
\label{eq:rbar}
\end{equation}
This is equivalent to applying a low-pass filter to the reward with time-constant $\flatfrac{1}{\rho}$.
We also refer to the difference $\delta_r(t) = r(t) - \bar{r}(t)$ as the reward prediction error (RPE), which can be interpreted as a global dopaminergic neuromodulatory signal, essential for learning in biological systems~\cite{schultz1997neural}.

\subsection{Learning}

Often, learning is viewed as separate from inference. Here, in contrast, we cast learning and inference as processes that co-evolve over time by making the parameters part of the system dynamics. That is, in Equations~\eqref{eq:sssm1} and \eqref{eq:sssm2}, we assume that $\vb*{\theta}(t)$ evolves over time in parallel to the other variables. In this sense, the only distinction between learning and inference is that the former is assumed to evolve at a slower time scale compared to the latter. 

The question remains how to set up learning dynamics such that the parameters adapt towards a more desirable state. To this end, we define learning as a stochastic process evolving forward in time. Specifically, let us assume that parameter dynamics are given by an Ornstein-Uhlenbeck process
\begin{equation}
\dd{\vb*{\theta}(t)} = \lambda (\vb*{\mu}(t) - \vb*{\theta}(t)) \dd{t} + \vb*{\Sigma} \dd{\vb*{W}(t)}
\label{eq:theta}
\end{equation}
with $\vb*{\mu}(t)$ the mean parameter, $\lambda$ the rate parameter, $\vb*{\Sigma} = \text{diag}(\sigma_1,\ldots, \sigma_n)$ the diffusion matrix, and $\vb*{W}(t) = (W_1(t),\ldots, W_n(t))^\top$ a stochastic process, which we take here to be a multivariate Wiener process. The Wiener process introduces stochastic perturbations, characterized by normally distributed increments with zero mean and variance proportional to $\dd{t}$, providing a mathematical model of random noise.

The OU process balances two key forces: (i) stability through mean-reversion via the term $\lambda (\vb*{\mu}(t) - \vb*{\theta}(t))$ and (ii) exploration through stochastic noise as the term $\vb*{\Sigma} \dd{\vb*{W}(t)}$ introduces randomness, allowing the system to explore the parameter space. Together, these terms embody the classical exploration-exploitation dilemma at the level of individual parameters. Exploration is driven by stochastic perturbations, which allow the parameters to sample a broad range of values. Exploitation, on the other hand, is guided by the mean-reverting force that nudges the parameters toward favorable regions identified by the current estimate of $\vb*{\mu}(t)$.

If we were to run \eqref{eq:theta} in isolation, the parameters would simply fluctuate around the mean $\vb*{\mu}(t)$, with no directed learning. To enable adaptation, we define the mean parameter dynamics using an ordinary differential equation:
\begin{equation}
\dd{\vb*{\mu}(t)} = \eta \delta_r(t) (\vb*{\theta}(t) -  \vb*{\mu}(t)) \dd{t}
\label{eq:mu}
\end{equation}
where $\eta$ is the learning rate, and $\delta_{r}(t)$ is the RPE, acting as a global modulatory reinforcement signal. This reinforcement mechanism dynamically adjusts $\vb*{\mu}(t)$, shifting the focus of exploration towards regions associated with higher reward. The term $(\vb*{\theta}(t) - \vb*{\mu}(t))$ ensures that the adaptation of $\vb*{\mu}$ reflects the influence of recent stochastic updates to $\vb*{\theta}$.
The interplay between the stochastic term and the mean-reversion term results in a probabilistic convergence behavior. As $\vb*{\mu}$ is refined through updates driven by $\delta_r$, the system converges to parameter values $\vb*{\theta}(T) \sim \mathcal{N}(\vb*{\mu}(T), \vb*{C})$ with mean $\vb*{\mu}(T)$ and stationary covariance $\vb*{C} = \flatfrac{\vb*{\Sigma} \vb*{\Sigma}^\top}{2 \lambda}$.

Since \eqref{eq:theta} defines learning dynamics in terms of an Ornstein-Uhlenbeck process, we refer to our proposed learning mechanism as Ornstein-Uhlenbeck adaptation (OUA). OUA combines stochastic exploration with adaptive exploitation, making it particularly well-suited for continual learning in dynamic, time-evolving environments, such as those encountered in neuromorphic systems.

\subsection{Experimental validation}

To test OUA as a learning mechanism, we designed experiments across several distinct scenarios. 
First, we analyzed the learning dynamics using a single-parameter model to gain insight into fundamental behavior. Subsequently, we examined recurrent and multi-parameter models to explore interactions among parameters and assess scalability. We then applied OUA to a real-world weather prediction task, forecasting temperature 24 hours ahead based on current measurements of temperature, humidity, wind speed, wind direction (expressed as sine and cosine components), and atmospheric pressure. The dataset used in this task contains hourly recordings for Szeged, Hungary, collected between 2006 and 2016.\footnote{Data can be obtained from \url{https://www.kaggle.com/datasets/budincsevity/szeged-weather/}.} Outliers were removed using linear interpolation and data was either standardized or whitened prior to further processing.
To further assess OUA, we tackled a control problem known as the stochastic double integrator (SDI), where the objective was to maintain a particle’s position and velocity near zero despite the effects of Brownian motion. We refer to this control task as the stochastic double integrator (SDI) problem~\cite{rao2001naive}. Finally, we extended our investigation to meta-learning, testing OUA’s ability to adapt hyper-parameters dynamically.

To implement OUA, we rely on numerical integration. The learning process is governed by a coupled system of stochastic and ordinary differential equations, defined in Equations~\eqref{eq:sssm1}, \eqref{eq:sssm2}, \eqref{eq:rbar}, \eqref{eq:theta}, and \eqref{eq:mu}. Specifically, Equations~\eqref{eq:sssm1} and \eqref{eq:sssm2} describe inference, while Equations~\eqref{eq:rbar}-\eqref{eq:mu} capture the learning dynamics. To integrate the system from the initial time $t_0$ to the time horizon $T$, we used an Euler-Heun solver implemented in the Python \texttt{Diffrax} package~\cite{kidger2021on}. In each experiment, the step size for numerical integration was set to $\Delta t = 0.05$. To interpolate inputs for the weather prediction task across the time window of interest, we used cubic Hermite splines with backward differences~\cite{morrill2021neural}. To ensure reproducibility, all scripts needed to replicate the results presented in this study are available via \url{https://github.com/artcogsys/OUA}.

\section{Results}

In the following, we demonstrate OUA-based learning in increasingly complex systems. Here, we suppress the time index $t$ from our notation to reduce clutter.

\subsection{Learning a single-parameter model}

\begin{figure}[!ht]
\centering
\includegraphics[width=\textwidth]{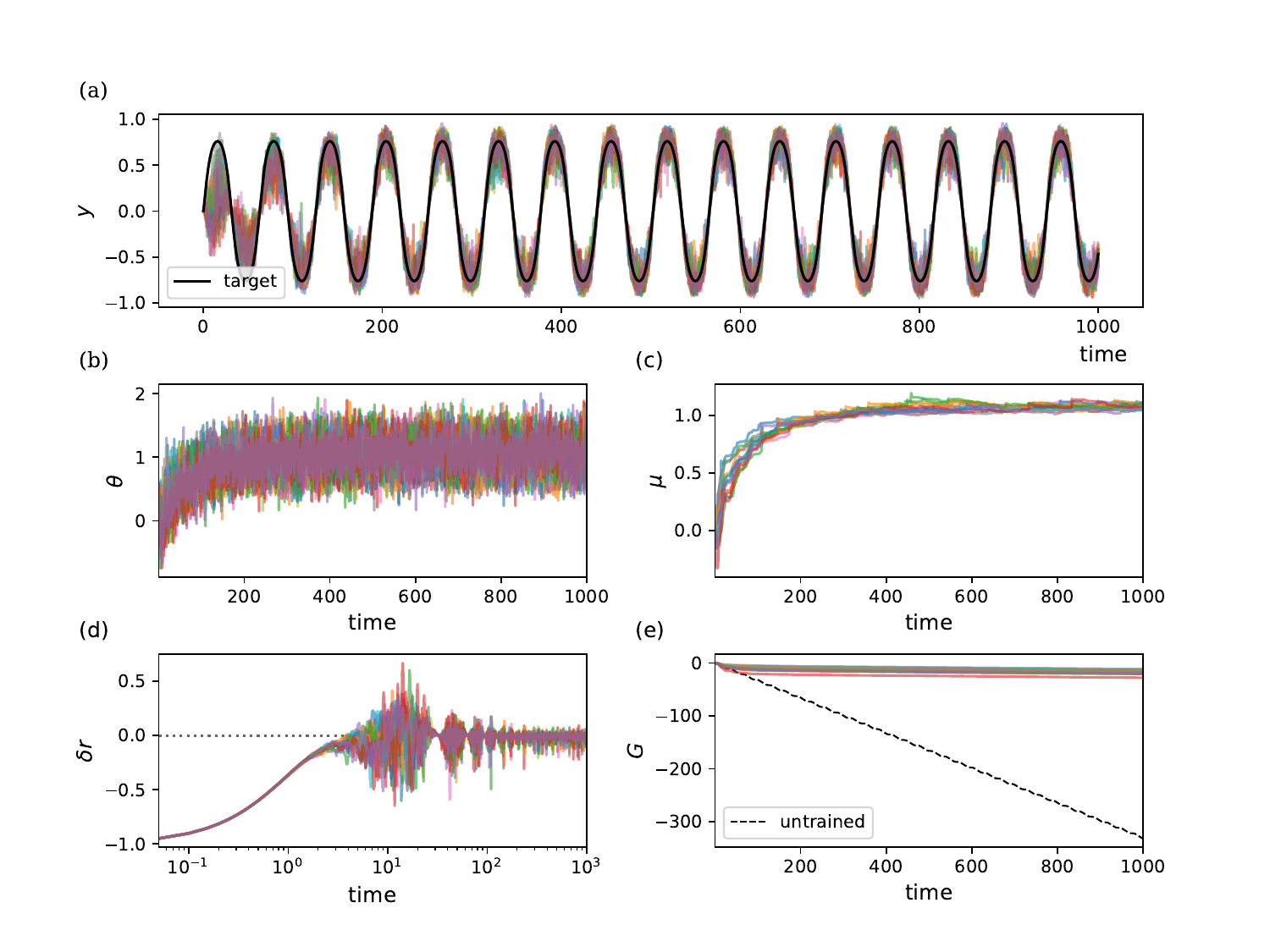}
\caption{Dynamics of a single-parameter model across 15 random seeds with $\rho = \lambda = \eta = 1$ and $\sigma = 0.3$. Initial conditions are $\bar{r}_0 = -1$, $\theta_0 = 0$, and $\mu_0 = 0$. The target output is generated with a fixed target parameter $\theta^* = 1$. (a) Target output vs model output (b) Evolution of $\theta$. (c) Trajectories of $\mu$. (d) RPE $\delta_r$, shown on a logarithmic axis to better visualise initial convergence. A dotted line at 0 is added to depict convergence around this value. (e) Cumulative reward $G$ over time, showing improvement with learning compared to the untrained model (dashed).}
\label{fig:one_parameter_learning}
\end{figure}

To analyze learning dynamics, we begin with a non-linear model containing a single learnable parameter $\theta$. We assume that
\begin{equation}
y = g_\theta(x) = \tanh(\theta x)
\end{equation}
with input $x$ and output $y$, and no latent state $z$ is used. The input is given by a sinusoidal signal $x(t) = \sin(0.1 \, t)$ and the reward is given by $r = -(y - y^*)^2$. The target output is given by $y^* = \tanh(\theta^* x)$, which is generated with a fixed parameter $\theta^* = 1$. Thus, this setup focuses on supervised learning for a non-linear input-output mapping. The learning dynamics for the single-parameter model are described by the following stochastic differential equations
\begin{align}
\dd{{\theta}} 
&= \lambda ({\mu} - {\theta}) \dd{t} + \sigma \dd{{W}}
\label{eq:multidynamics1}
\\
\dd{{\mu}} 
&= \eta \delta_r ({\theta} - \mu)  \dd{t}
\label{eq:multidynamics2}
\end{align}
where $\delta_r = r - \bar{r}$, $\bar{r}$ is the expected reward, and $W$ is a standard Wiener process. Here, $\theta$ represents the model parameter, and $\mu$ represents its estimate, with the equations incorporating noise through $\dd{W}$.  


Figure~\ref{fig:one_parameter_learning} illustrates the learning dynamics of this model, simulated over 15 trials using different random seeds. 

Figure~\ref{fig:one_parameter_learning}a depicts the target output vs the model output for the different random seeds. In Figure~\ref{fig:one_parameter_learning}b and~\ref{fig:one_parameter_learning}c, we observe the trajectories of $\theta$ and $\mu$ as they converge towards values that allow the model to approximate the target output. Due to the stochastic nature of the dynamics, convergence exhibits probabilistic behavior, leading to variability in $\theta$ (and thus $\mu$) around their optimal values. Nonetheless, OUA ensures the model’s output closely follows the target, even in the presence of continuous noise.  

Figure~\ref{fig:one_parameter_learning}d shows how the reward prediction error (RPE), $\delta_r$, tends to 0 over time. Despite achieving convergence, $\delta_r$ remains noisy due to the intrinsic stochasticity in the parameters. Figure~\ref{fig:one_parameter_learning}e displays the accumulative reward ($G$) over time. The dashed line represents the return for an untrained model ($\theta = \theta_0$). Results demonstrate that learning significantly improves return, with variability in the accumulative reward arising from differences in the time it takes for $\theta$ to converge in individual trials.

\begin{figure}[!ht]
\centering
\includegraphics[width=\textwidth]{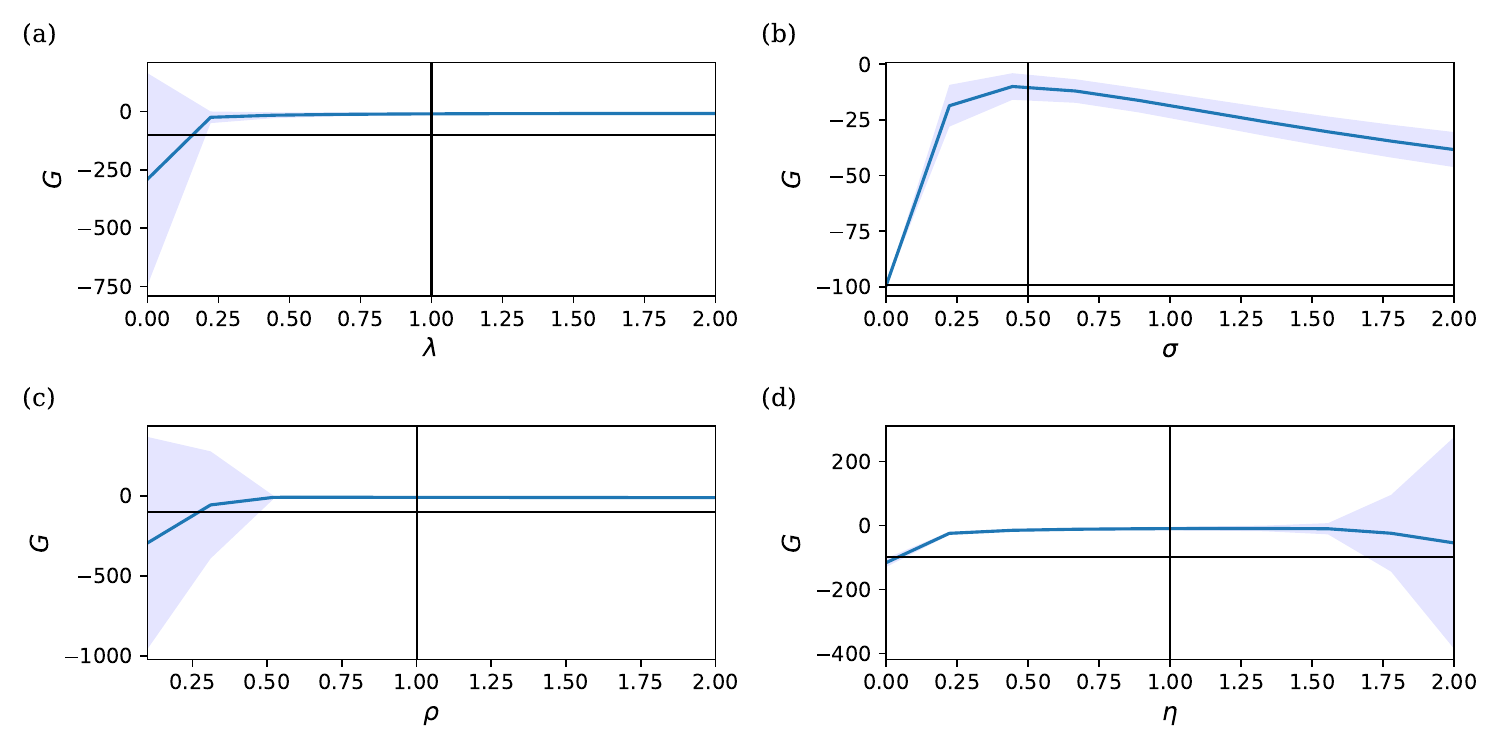}
\caption{Sensitivity of the final cumulative reward $G(T)$ to model hyper-parameters for the input-output mapping task. Results are averaged over 15 runs using different random seeds. The shaded area represents variability across runs, showing stability across a wide range of hyper-parameter settings. Vertical lines show the chosen hyper-parameter values, and horizontal lines show the return without learning. (a) Impact of $\lambda$. (b) Impact of $\sigma$. (c) Impact of $\rho$. (d) Impact of $\eta$.}
\label{fig:parameter_sensitivity}
\end{figure}

Analyzing the sensitivity of parameter convergence to hyper-parameter choices provides valuable insights. Figure~\ref{fig:parameter_sensitivity} shows how the choice of hyper-parameters influences the final obtained cumulative reward for the same task (input-output mapping learning). For all hyper-parameters, we see a clear peak in the return $G$, except for $\rho$ since the true average reward is equal to the initial estimate of $\bar{r}_0 = 0$. Even for high noise levels $\sigma$, we still observe effective learning.

\subsection{Learning in recurrent systems}

While the previous analysis explored learning a static input-output mapping, we now examine the learning dynamics of a non-linear recurrent system, given by
\begin{equation}
\begin{split}
\dd{{z}} 
&= (f( \theta_1 z +  \theta_2 x ) - z) \dd{t} \\
{y} &= \theta_3 z 
\end{split}
\label{eq:CTRNN}
\end{equation}
where $x$ is the input, $z$ is the latent state and $y$ is the output. This model can be interpreted as a continuous-time recurrent neural network (CTRNN) or a latent ordinary differential equation (ODE).
The reward is given by $r = -(y - y^*)^2$, where $y^*$ is the target output generated using fixed target parameters $\vb*{\theta}^* = (\theta_1, \theta_2, \theta_3) = (0.3, 0.7, 1.0)$. This setup requires fast dynamics in $z$ and slow dynamics in $\vb*{\theta}$, creating a challenging learning scenario.  We demonstrate OUA's capability to learn the parameters of a recurrent model by training it on a 1D input-output mapping.

\begin{figure}[!h]
\centering
\includegraphics[width=\textwidth]{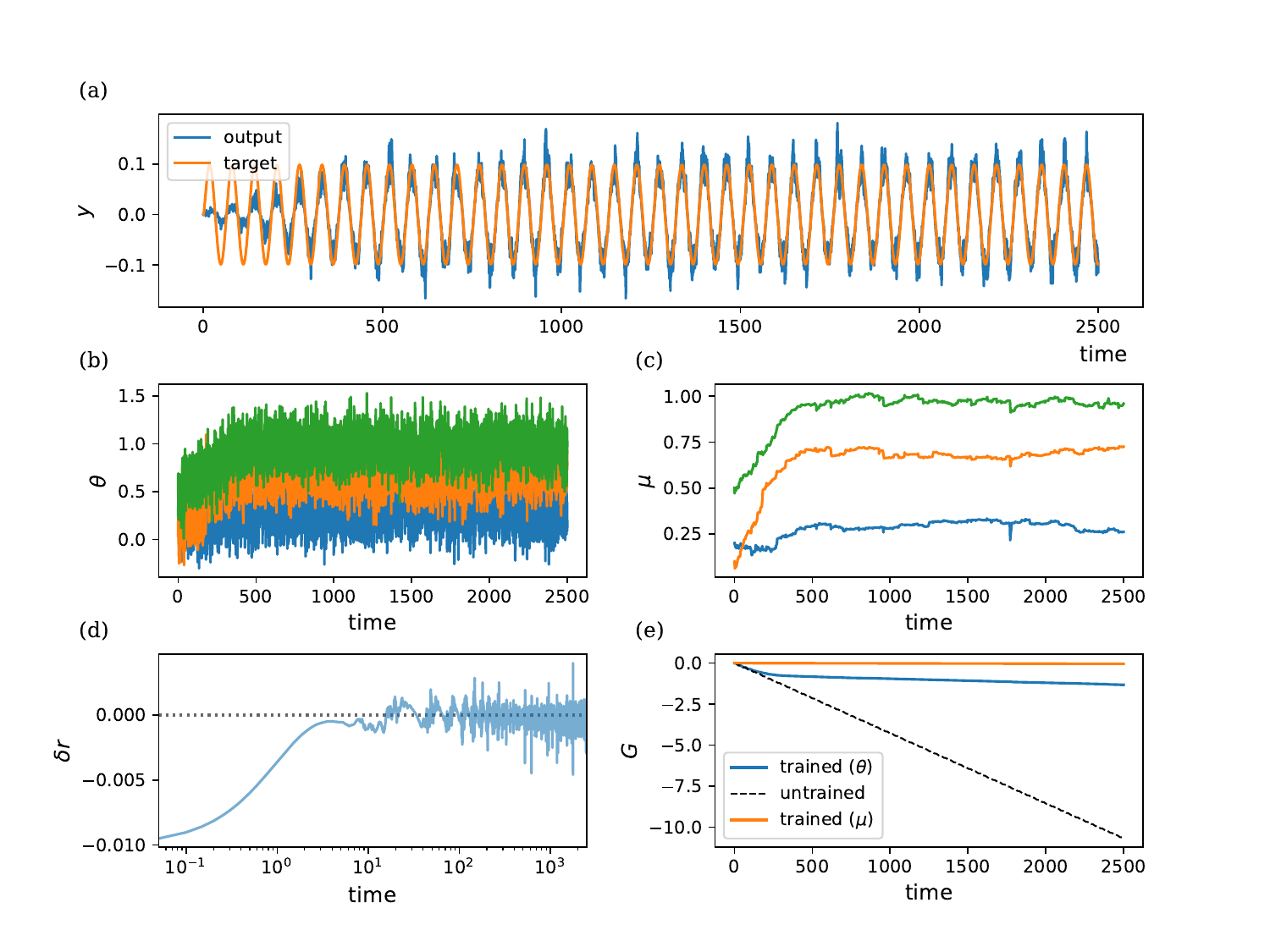}
\caption{
Learning dynamics in a recurrent model with parameters $\vb*{\theta} = [\theta_1, \theta_2, \theta_3]$, where $\rho=\lambda=1$, $\eta=50$ and $\sigma=0.2$. Initial conditions are $\bar{r}_0 = -0.1$, $\vb*{\theta_{0}}=[0.2, 0.1, 0.5]$. The fixed target parameters used to generate the target output are $\vb*{\theta}^*=[0.3, 0.7, 1.0]$. (a) Target output vs model output (b) Evolution of the parameters $\vb*{\theta}$. (c) Trajectories of $\vb*{\mu}$. (d) RPE $\delta_r = r - \bar{r}$, shown on a logarithmic axis to better visualise initial convergence. A dotted line at 0 is added to depict convergence around this value. (e) Cumulative reward $G$ over time, showing improvement with learning compared to the untrained model (dashed). The blue line indicates the return during parameter learning. The dashed line denotes the return when $\vb*{\theta}$ are fixed to their initial values $\vb*{\theta}_0$. The orange line indicates the return obtained when we fix parameters to the final mean parameters $\vb*{\theta}(t) = \vb*{\mu}(T)$.
}
\label{fig:1D_recurrent}
\end{figure}

Figure~\ref{fig:1D_recurrent} illustrates the learning dynamics of this three-parameter recurrent system, which includes connections from the input to the latent state, recurrent dynamics within the latent state, and connections from the latent state to the output. 

Figure~\ref{fig:1D_recurrent}a depicts the target output vs the model output for the different random seeds. In Figure~\ref{fig:1D_recurrent}b and~\ref{fig:1D_recurrent}c, we observe the trajectories of $\theta$ and $\mu$ as they converge towards values that allow the model to approximate the target output. We can observe the same probabilistic convergence behavior described in the non-recurrent model. Figure~\ref{fig:one_parameter_learning}d shows how the reward prediction error (RPE), $\delta_r$, tends to 0 over time. As in the non-recurrent model, $\delta_r$ remains noisy after convergence due to the intrinsic stochasticity in the parameters. Figure~\ref{fig:one_parameter_learning}e displays the accumulative reward ($G$) over time. The dashed line represents the return for an untrained model.

\subsection{Learning a multi-parameter model}

\begin{figure}[!h]
\centering
\includegraphics[width=\textwidth]{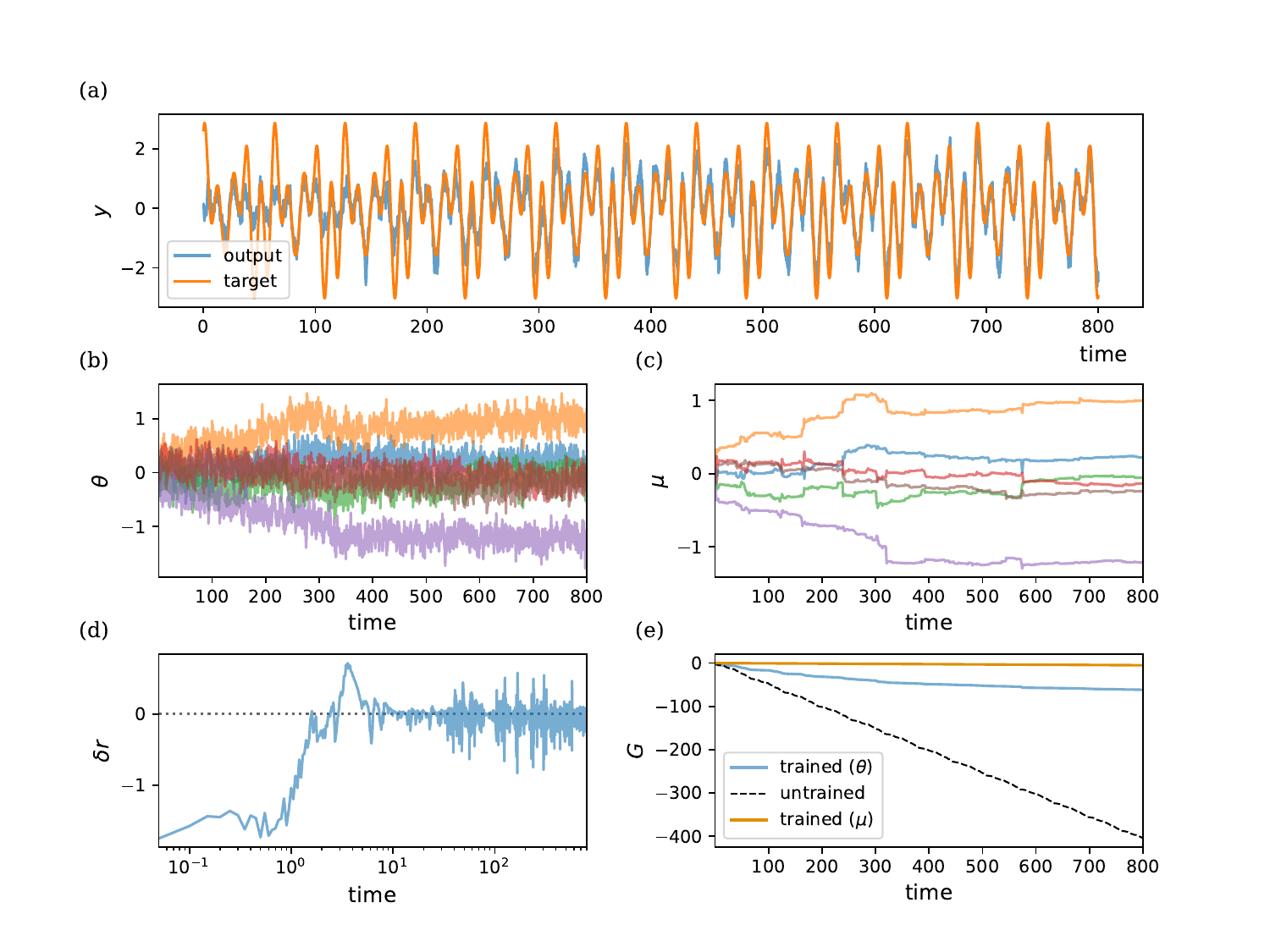}
\caption{
Learning dynamics in a model with parameters $\vb*{\theta} = (\theta_1, \ldots, \theta_6)^\top$, where $\rho = \lambda = \eta = 1$ and $\sigma = 0.2$. Initial conditions are $\bar{r}_0 = -1$ and $\vb*{\theta}_0 = \vb*{\mu}_0 = \mathbf{0}$. The fixed target parameters used to generate the target output are $\vb*{\theta}^* = (0.3, 1.1, 0.0, -0.3, -1.5, -0.4)^\top$. (a) Target output vs model output. (b) Evolution of the parameters $\vb*{\theta}$. (c) Trajectories of $\vb*{\mu}$. (d) RPE $\delta_r = r - \bar{r}$, shown on a logarithmic axis to better visualise initial convergence. A dotted line at 0 is added to depict convergence around this value. (e) Cumulative reward $G$ over time, showing improvement with learning compared to the untrained model (dashed). The blue line indicates the return during parameter learning. The dashed line denotes the return when $\vb*{\theta}$ are fixed to their initial values $\vb*{\theta}_0$. The orange line indicates the return obtained when we fix parameters to the final mean parameters $\vb*{\theta}(t) = \vb*{\mu}(T)$.}
\label{fig:multivariate_input}
\end{figure}

Having demonstrated the feasibility of learning with a single parameter, we now investigate whether effective learning extends to cases involving multiple parameters $\vb*{\theta} = (\theta_1,\ldots, \theta_n)^\top$. For this, we assume a model given by
\begin{equation}
y = g_{\vb*{\theta}}(\vb*{x}) = \tanh(\vb*{\theta}^\top \vb*{x})
\label{eq:multiparam}
\end{equation}
where $\vb*{x} = (x_1,\ldots,x_n)^\top$ is the input vector and $y$ is the scalar output. The input $\vb*{x}$ is composed of multiple sine waves, defined as $x_i(t) = \sin(i , 0.1 , t + (i-1) 2\pi/n)$ for $1 \leq i \leq n$. The reward is given by $r = -(y - y^*)^2$ with target output $y^* = \tanh((\vb{\theta}^*)^\top \vb*{x})$, generated with fixed target parameters $\vb*{\theta}^* = (0.3, 1.1, 0.0, -0.3, -1.5, -0.4)^\top$. Finally, learning dynamics are given by
\begin{align}
\dd{\vb*{\theta}} 
&= \lambda (\vb*{\mu} - \vb*{\theta}) \dd{t} + \vb*{\Sigma} \dd{\vb*{W}}\\
\dd{\vb*{\mu}} 
&= \eta \delta_r (\vb*{\theta} - \vb*{\mu})  \dd{t}
\end{align}
with $\vb*{\Sigma} = \sigma \vb*{I}$ and $\vb*{W} = (W_1,\ldots, W_n)^\top$ a multivariate standard Wiener process.

Figure~\ref{fig:multivariate_input} shows that effective learning can still be achieved when having multiple parameters. Figure~\ref{fig:multivariate_input}a depicts the target output vs the model output. In Figure~\ref{fig:multivariate_input}b and~\ref{fig:multivariate_input}c, we observe the trajectories of $\theta$ and $\mu$ as they converge towards values that allow the model to approximate the target output. Figure~\ref{fig:multivariate_input}d shows how the reward prediction error (RPE), $\delta_r$, tends to 0 over time. As in the previous models, $\delta_r$ remains noisy after convergence due to the intrinsic stochasticity in the parameters. Figure~\ref{fig:multivariate_input}e displays the accumulative reward ($G$) over time. The dashed line represents the return for an untrained model.


Note that we may also choose to use the final mean $\vb*{\mu}(T)$ as the parameters estimated after learning. While this steps away from the continual learning setting, it is of importance when deploying trained systems in real-world applications. The orange line in Fig.~\ref{fig:multivariate_input}d shows that this indeed provides optimal performance.

\subsection{Weather prediction task}

We now apply our approach to a real-world weather prediction task. The input vector $\vb*{x}$ consists of several weather features: current temperature, humidity, wind speed, sine of the wind direction angle, cosine of the wind direction angle, and another humidity measurement. The model aims to predict the temperature, denoted as $y$, 24 hours ahead. The setup follows the structure used in the multiple-parameter analysis, with the main difference being that $y^*$ now represents the target temperature 24 hours ahead. The predicted temperature is given by $y = \vb*{\theta}^\top \vb*{x}$. 
To validate test performance, a separate segment of the weather dataset was used. 
Additionally, motivated by recent work which shows that input decorrelation improves learning efficiency~\cite{ahmad2022constrained, ahmad2024correlations}, we tested the model both with and without applying ZCA whitening~\cite{bell1997independent} to the input features.

\begin{figure}[!ht]
\centering
\includegraphics[width=\textwidth]{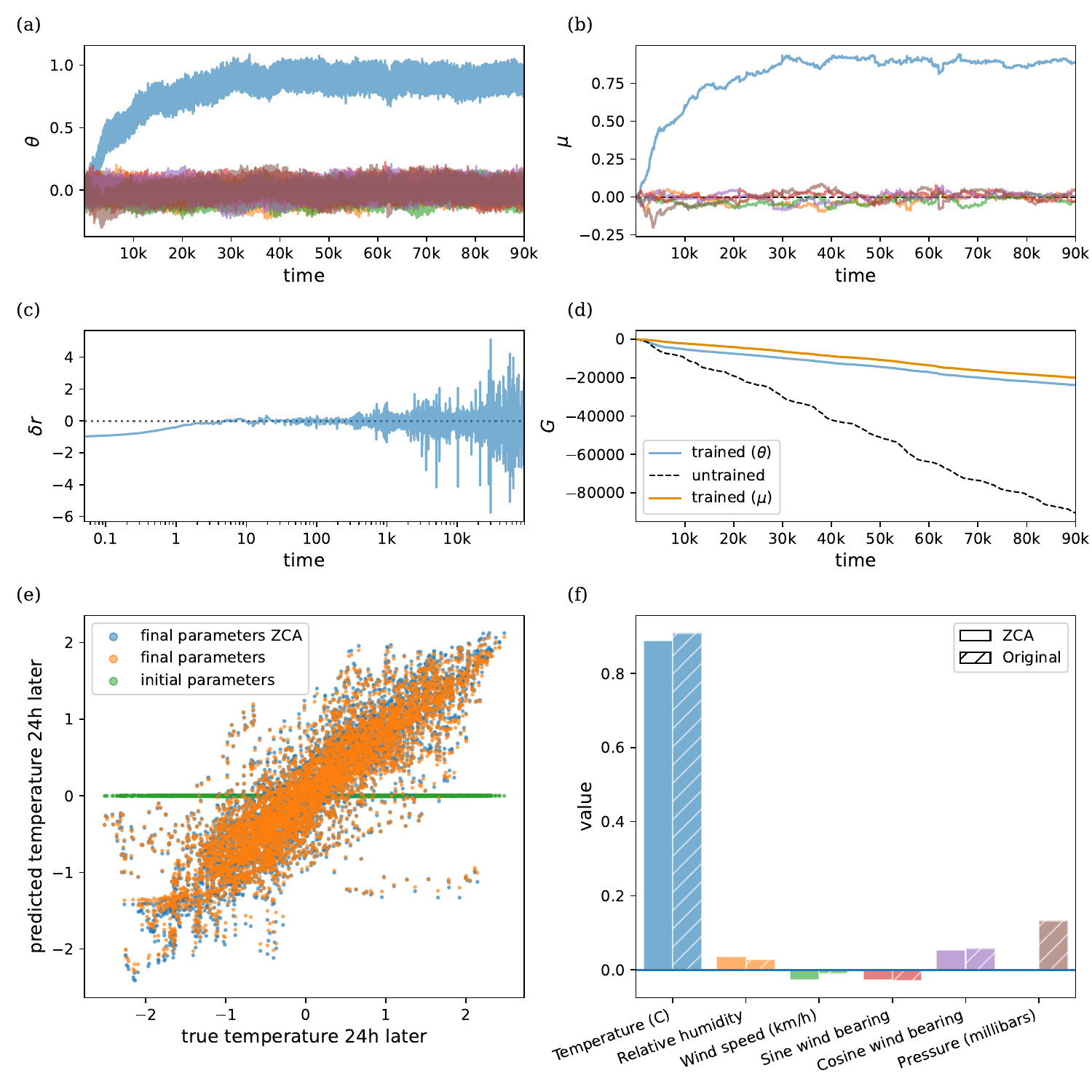}
\caption{
Dynamics of the weather prediction model whose parameters $\vb*{\theta} = (\theta_1,\ldots, \theta_6)$ denote the different weather features, respectively. The goal is to learn to predict the temperature 24 hours ahead. Hyper-parameters: $\rho=\lambda=1$, $\sigma=0.05$, $\eta=0.1$. Initial conditions: $\bar{r} = 0$, $\vb*{\theta}_0 \sim \mathcal{N}(\vb*{0}, 10^{-6} \vb*{I})$, $\vb*{\mu}_0 = \vb*{\theta}_0$. (a) Dynamics of parameters $\vb*{\theta}$. (b) Dynamics of means $\vb*{\mu}$. (c) RPE $\delta_r = r - \bar{r}$, shown on a logarithmic axis to better visualise initial convergence. A dotted line at 0 is added to depict convergence around this value. (d) Cumulative reward $G$ over time, showing improvement with learning compared to the untrained model (dashed). The blue line indicates the return during parameter learning. The dashed line denotes the return when $\vb*{\theta}$ are fixed to their initial values $\vb*{\theta}_0$. The orange line indicates the return obtained when we fix parameters to the final mean parameters $\vb*{\theta}(t) = \vb*{\mu}(T)$ (e) Predicted vs. true 24h-ahead temperature on test data, using final $\vb*{\mu}(T)$ with/without ZCA and initial $\vb*{\theta}_0$. (f) Final parameter values for each regressor. For whitened data, $\vb*{\mu}(T) \vb*{R}$ projects coefficients back into the original space.}
\label{fig:weather_combined}
\end{figure}

Figure~\ref{fig:weather_combined}a-c shows the learning dynamics for this task. Figure~\ref{fig:weather_combined}d shows that, during training, the model learns to improve the cumulative reward. Figure~\ref{fig:weather_combined}e shows a scatter plot comparing the true versus predicted 24-hour ahead temperature before training (green) and after training, either with ZCA (blue) or without ZCA (orange). The final mean values $\vb*{\mu}(T)$ were used as the parameters $\vb*{\theta}(t)$ when performing inference on separate test data. Results show that accurate prediction is achieved with a Pearson correlation of 0.871, between true and predicted temperatures, both with and without ZCA, respectively. Furthermore, a mean squared error (MSE) of 0.264 and 0.316 was achieved with and without ZCA, respectively.
Figure~\ref{fig:weather_combined}f shows these final mean values, indicating that the current temperature mostly determines the prediction outcome.

\subsection{Learning to control a stochastic double integrator}

Next, we explore the task of controlling a stochastic double integrator (SDI)~\cite{rao2001naive}, which presents a more complex learning environment compared to the supervised learning tasks considered earlier. In this setup, the agent's actions influence the controlled system, which in turn affects the agent's observations. This feedback loop can lead to potentially unstable dynamics if not handled properly.

Let $\vb*{s} = (s_1, s_2)^\top$ represent the state vector, where $s_1$ and $s_2$ denote the position and velocity of a particle moving in one dimension. The SDI system is described by the following state-space equations
\begin{align}
\dd{\vb*{s}} 
&= \left(\begin{bmatrix}
0 & 1 \\
0 & -\gamma 
\end{bmatrix} \vb*{s} + \begin{bmatrix}
0\\
1
\end{bmatrix} y \right) \dd{t} + \begin{bmatrix}
0\\
\alpha
\end{bmatrix}\dd{W}\label{eq:sssm1b}
\\
\vb*{x} &= \vb*{s} + \beta \vb*{\epsilon}
\label{eq:sssm2b}
\end{align}
where $\vb*{\epsilon} \sim \mathcal{N}(\vb*{0}, \vb*{I})$. Here, $\alpha$ represents process noise, $\beta$ represents observation noise and $\gamma$ represents a friction term. Note that we again employ an OU process to model the stochastic dynamics of the velocity. 
The reward is given by a (negative) quadratic cost $r= -0.5  ||\vb*{s}||^2 - 0.5 y^2$, penalizing deviations of the state $\vb*{s}$ from the set-point, where both the position and velocity are equal to zero, while also penalizing large values of the control $y$. The agent is defined by $y = \vb*{\theta}^\top \vb*{x}$, with learning dynamics given by Equations~\eqref{eq:multidynamics1} and \eqref{eq:multidynamics2}, as before. Note that the output $y$ of the agent is the control input to the SDI whereas the SDI generates the observations $\vb*{x}$ to the agent. Hence, both the agent and the environment are modeled as coupled dynamical systems.

\begin{figure}[!ht]
\centering
\includegraphics[width=\textwidth]{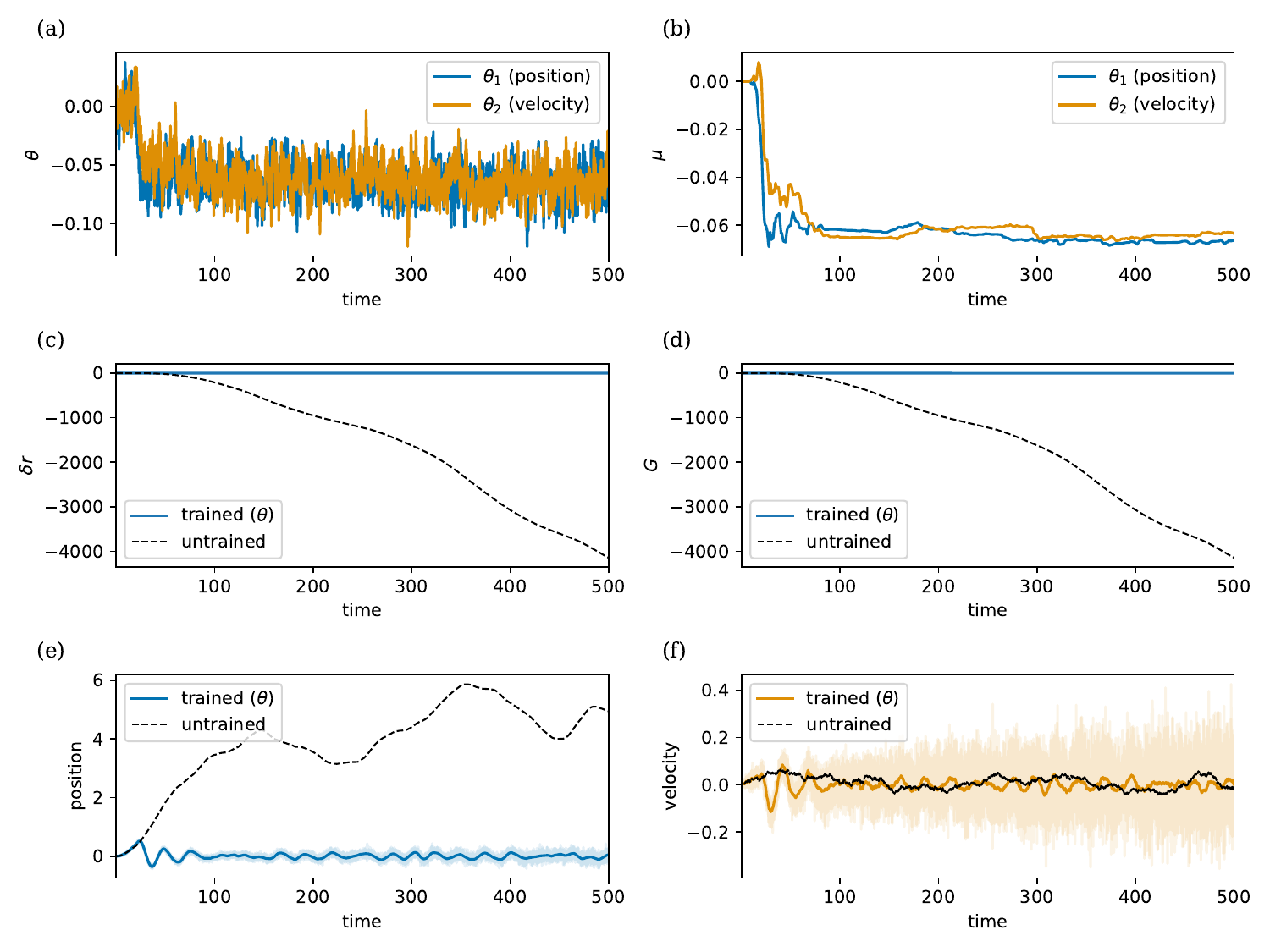}
\caption{
Learning to control a stochastic double integrator. Hyper-parameters: $\rho=2$, $\lambda=1$, $\sigma=0.02$, $\eta=50$, $\gamma = 0.01$, $\alpha=\beta=0.005$. Initial conditions are set to zero for all variables. (a) Dynamics of parameters $\vb*{\theta}$. (b) Dynamics of means $\vb*{\mu}$. (c) Reward prediction error $\delta_r = r - \bar{r}$. (d) Return $G$ over time, where higher is better. Blue line: return during learning; dashed line: return with fixed $\vb*{\theta}_0$. (e) Particle position after learning (blue) with observation noise (light blue); dashed line: position changes without learning. (f) Particle velocity after learning (orange) with observation noise (light orange); dashed line: velocity fluctuations without learning.
}
\label{fig:control_task}
\end{figure}

Figure~\ref{fig:control_task} shows that OUA can learn to control a stochastic double integrator, demonstrating effective learning in this more challenging control setting. Both the parameters for the position and the velocity converge to negative values. This is indeed optimal since it induces accelerations that move the particle's position and velocity to zero.

Figure~\ref{fig:control_task} illustrates the learning dynamics of the stochastic double integrator control task. As shown in Figure~\ref{fig:control_task}a-c, the agent learns to adjust the parameters $\vb*{\theta}$ and their means $\vb*{\mu}$ during the learning process. Figure~\ref{fig:control_task}d shows that the return improves over time, with the model achieving a better performance compared to the baseline, where $\vb*{\theta}$ is fixed to its initial values $\vb*{\theta}_0$.

The learning process allows the agent to effectively control the particle, as seen in Figures~\ref{fig:control_task}e and~\ref{fig:control_task}f. The dashed lines represent the behavior of the particle without learning, showing a significant deviation of the position from zero. The learned controller, however, successfully adjusts the velocity to drive the position to zero, demonstrating that the agent has effectively learned to stabilize the system. Thus, OUA can successfully learn to control the stochastic double integrator, even in the presence of observation and process noise.

\subsection{Meta-learning}

In the previous analyses, we estimated the parameters $\vb*{\theta}$ and $\vb*{\mu}$ while keeping the hyper-parameters fixed. However, we can also choose to learn the hyper-parameters using the same mechanism. To illustrate this, we introduce a learnable diffusion coefficient $\sigma$ for the single-parameter model. Rather than fixing $\sigma$, we allow it to adjust, which in turn modulates the exploration versus exploitation trade-off in the learning process. This can be viewed as a form of meta-learning, where the adjustment of $\sigma$ influences the dynamics of learning based on the problem at hand~\cite{Schmidhuber1987}.

\begin{figure}[!ht]
\centering
\includegraphics[width=\textwidth]{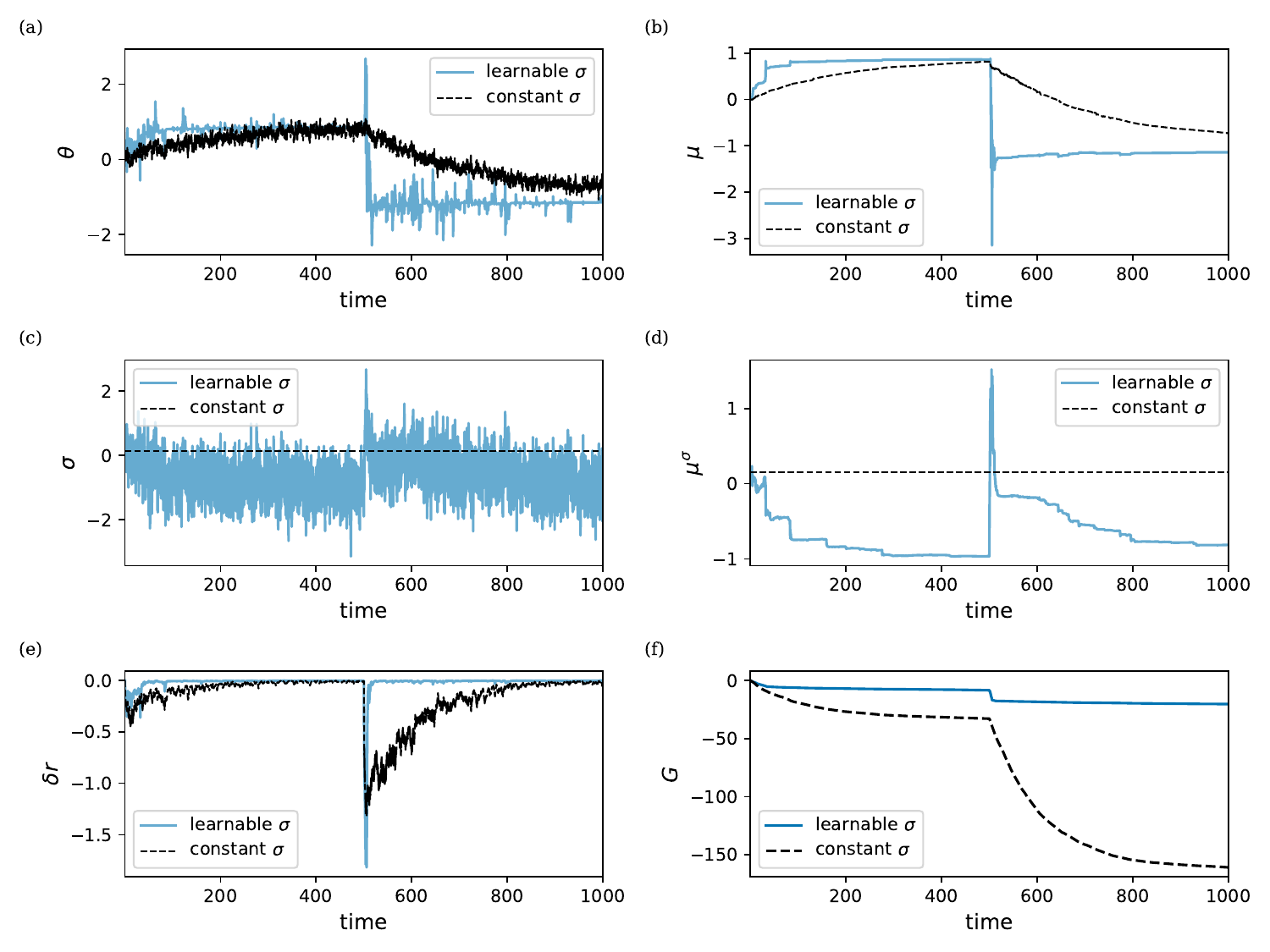}
\caption{Results when using a learnable $\sigma$ compared to a constant $\sigma = \sigma_0$. Parameters are set to $\rho = \lambda= \eta=1$, $\lambda^\sigma = 2$ and $\eta^\sigma = 3.0$ for the model with learnable $\sigma$. The initial conditions are given by $\bar{r} = \theta_0=\mu_0=0$ and $\sigma_0 = \mu^\sigma_0 = 0.15$. Dashed lines indicate results when using a constant $\sigma$. (a) Dynamics of $\theta$. (b) Dynamics of $\mu$. (c) Dynamics of $\sigma$. (d) Dynamics of $\mu^\sigma$. (e) Dynamics of $\bar{r}$. (f) Dynamics of $G$.}
\label{fig:meta_learning}
\end{figure}

To implement meta-learning, we define additional dynamics for $\sigma$, given by
\begin{align}
\dd{{\sigma}} 
&= \lambda^\sigma (\mu^\sigma - {\sigma}) \dd{t} + \rho \dd{W} \\ 
\dd{\mu^\sigma} 
&= \eta^\sigma \delta_r (\sigma -\mu^\sigma) \dd{t}
\end{align}
where $W$ is a standard Wiener process, analogous to the equations for $\theta$ and $\mu$ in previous sections (Equations~\eqref{eq:multidynamics1} and \eqref{eq:multidynamics2}).

To verify that the meta-learning process effectively identifies the optimal value for the chosen hyper-parameter, in this case, $\sigma$, we test the model in a volatile environment. For this, we tackled a simple input-output mapping task, where the target parameter $\theta^*$, used to generate the target output, switches from +1 to -1 during learning.


Figure~\ref{fig:meta_learning} presents the learning dynamics of $\sigma$ under these volatile conditions. As seen in Figure~\ref{fig:meta_learning}a, the system converges faster to the optimal $\theta$ when meta-learning is employed, compared to using a fixed $\sigma$. Additionally, Figure~\ref{fig:meta_learning} demonstrates how $\sigma$ adapts by increasing its value when $\theta$ changes, promoting exploration, and decreasing when the target remains stable, favoring exploitation. The faster convergence is also reflected by the larger return in Fig.~\ref{fig:meta_learning}f. Figures~\ref{fig:meta_learning}c and \ref{fig:meta_learning}d show that $\sigma$ and $\mu^\sigma$ rapidly adapt to a more suitable higher value of $\sigma$, encouraging fast exploration. At a later stage, we observe adaptation to a value of $\sigma$ below the initial $\sigma_0$ value, encouraging exploitation.

\section{Discussion}

In this paper, we introduced Ornstein-Uhlenbeck adaptation as a learning mechanism for naturally and artificially intelligent systems. Our results show that learning can emerge purely from simulating the dynamics of parameters governed by an Ornstein-Uhlenbeck process, where mean values are updated based on a global reward prediction error. We showed that OUA effectively learns in supervised and reinforcement learning settings, including single- and multi-parameter models, recurrent systems, and meta-learning tasks. Notably, meta-learning via OUA allows the model to balance exploration and exploitation~\cite{Feldbaum1960,Sutton2015} automatically. We hypothesize that, in this setting, the remaining stochastic drift of $\vb*{\theta}$ around its mean approximates the posterior distribution of $\vb*{\theta}$, similar to Bayesian methods~\cite{Xu2022}.

OUA offers significant advantages for machine learning as it offers a gradient-free learning framework. Unlike backpropagation, no exact gradients, backward passes, or non-local information beyond access to a global reward signal is needed. OUA is also highly parallelizable since multiple agents may simultaneously explore the parameter space using one global parameter mean $\vb*{\mu}$, which may have implications for federated learning~\cite{mcmahan2017communication}. Our experiments highlight the importance of hyper-parameter selection for learning stability (Figure~\ref{fig:parameter_sensitivity}), suggesting that meta-learning or black-box optimization techniques, such as Bayesian optimization~\cite{snoek2012practical, shahriari2015taking} or evolution strategies~\cite{beyer2002evolution}, could further enhance OUA’s performance.

It should be noted that OUA is inherently a stochastic algorithm since noise fluctuations drive learning. This means that, in the limit, instead of converging towards exact parameter values, individual parameter values ${\theta}$ fluctuate about their mean ${\mu}$ with variance $\flatfrac{\sigma^2 }{2\lambda}$. This is also reflected by induced RPE fluctuations around zero. We consider this behavior a feature rather than a bug since it allows the agent to adapt to changing circumstances by continuously probing for better parameter values. This can be seen in Fig.~\ref{fig:meta_learning}, where the system responds to volatile target parameters. As also shown in this figure, the meta-learning formulation further allows the system to reduce parameter variance in stable situations. Suppose one does want to enforce fixed parameter values, as shown in Figures~\ref{fig:multivariate_input} and \ref{fig:weather_combined}, one may choose to set the parameters $\vb*{\theta}$ to their estimated mean values $\vb*{\mu}$, or enforce convergence by increasing the rate parameter $\lambda$ (or decreasing the noise variance $\sigma^2$) over time, similar to the use of learning rate schedulers in conventional neural network training. 

A key area for future work is scaling OUA to more complex tasks. This could involve replacing Wiener processes with other noise processes, such as compound Poisson processes, to selectively update subsets of parameters, potentially reducing weight entanglement. Additionally, as demonstrated in Figure~\ref{fig:weather_combined}, decorrelating the input data via ZCA whitening accelerates convergence~\cite{ahmad2022constrained, ahmad2024correlations}. Such decorrelation strategies can generalize across deep and recurrent networks~\cite{dalm2024efficient, Fernandez2024}, and even be implemented locally as part of the forward dynamics~\cite{ahmad2024correlations}. OUA is also well-suited for training deep networks, which can be interpreted in our framework as recurrent networks with block-diagonal structures.

OUA’s implications extend to biological learning as well, providing insights into potential mechanisms for noise-based learning in the brain~\cite{Faisal2008}. While our framework is inspired by biological processes, it operates at a conceptual level rather than simulating specific neurophysiological mechanisms. For instance, we draw a connection with the stochastic release of neurotransmitters~\cite{branco2009probability, rolls2010noisy}, which introduces variability in synaptic transmission, akin to the exploratory noise in OUA’s parameter updates. This stochasticity, paired with global reward signals such as dopaminergic RPEs, could guide synaptic weight adjustments dynamically, as modeled by the mean-reverting updates in OUA. Additionally, OUA shares conceptual similarities with reward-modulated Hebbian learning, where the update of the synaptic weight $\theta_{ij}$ is a function of the pre- and post-synaptic activity (specifically, the activity deviation from their mean activity), and the RPEs~\cite{legenstein2010reward}. Formally, while RMHL provides a biologically plausible interpretation of node perturbation~\cite{hiratani2022stability, williams1992simple, werfel2003learning, fiete2006gradient, dalm2024efficient, Fernandez2024}, OUA offers a biologically plausible interpretation of weight perturbation~\cite{zuge2023weight, cauwenberghs1992fast, dembo1990model}. Importantly, the conceptual parallels we draw to brain processes serve to provide a high-level understanding of how noise and local reward signals might facilitate efficient learning in biological systems. We do not aim to model exact neurophysiological processes but to leverage these principles in a computational framework that remains biologically inspired.

Our work is also of particular relevance for neuromorphic computing and other unconventional (non-von-Neumann-style) architectures~\cite{Markovic2020,scellier2017equilibrium,Stern2021,Nakajima2022,lopez2023,Momeni2023,doremaele2024}, where learning and inference emerge from the physical dynamics of the system~\cite{Jaeger2023,Kaspar2021}. These approaches pave the way for sustainable, energy-efficient intelligent systems. Unlike conventional AI, where learning and inference are treated as distinct processes, our method integrates them seamlessly. It relies solely on online adaptation of parameter values through drift and diffusion terms, which can be directly implemented in physical systems. OUA’s reliance on local, continuous-time parameter updates makes it a natural fit for spiking neuromorphic systems~\cite{Shahsavari2023}, eliminating the need for differentiability and backpropagation.

Looking ahead, we envision future AI systems where intelligence emerges purely from running a system’s equations of motion forward in time to maximize efficiency and effectiveness. OUA exemplifies how such physical learning machines could be realized through local operations, leveraging the mean-reverting dynamics of the Ornstein-Uhlenbeck process. Future work will focus on theoretical extensions, scaling OUA to high-dimensional tasks, addressing challenges like delayed rewards and catastrophic forgetting, and implementing OUA on neuromorphic and unconventional computing platforms.

\section*{Acknowlegments}
This publication is part of the DBI2 project (024.005.022, Gravitation), which is financed by the Dutch Ministry of Education (OCW) via the Dutch Research Council (NWO).

\bibliographystyle{plain}
\bibliography{references}

\begin{thebibliography}{10}

\bibitem{ahmad2024correlations}
Nasir Ahmad.
\newblock Correlations are ruining your gradient descent.
\newblock {\em ArXiv preprint arXiv:2407.10780}, 2024.

\bibitem{ahmad2022constrained}
Nasir Ahmad, Ellen Schrader, and Marcel van Gerven.
\newblock Constrained parameter inference as a principle for learning.
\newblock {\em ArXiv preprint arXiv:2203.13203}, 2022.

\bibitem{bell1997independent}
Anthony~J Bell and Terrence~J Sejnowski.
\newblock The “independent components” of natural scenes are edge filters.
\newblock {\em Vision Research}, 37(23):3327--3338, 1997.

\bibitem{bellec2020solution}
Guillaume Bellec, Franz Scherr, Anand Subramoney, Elias Hajek, Darjan Salaj, Robert Legenstein, and Wolfgang Maass.
\newblock A solution to the learning dilemma for recurrent networks of spiking neurons.
\newblock {\em Nature Communications}, 11(1):3625, 2020.

\bibitem{bengio2014auto}
Yoshua Bengio.
\newblock How auto-encoders could provide credit assignment in deep networks via target propagation.
\newblock {\em ArXiv preprint arXiv:1407.7906}, 2014.

\bibitem{beyer2002evolution}
Hans-Georg Beyer and Hans-Paul Schwefel.
\newblock Evolution strategies -- a comprehensive introduction.
\newblock {\em Natural Computing}, 1:3--52, 2002.

\bibitem{bienenstock1982theory}
Elie~L Bienenstock, Leon~N Cooper, and Paul~W Munro.
\newblock Theory for the development of neuron selectivity: orientation specificity and binocular interaction in visual cortex.
\newblock {\em Journal of Neuroscience}, 2(1):32--48, 1982.

\bibitem{branco2009probability}
Tiago Branco and Kevin Staras.
\newblock The probability of neurotransmitter release: variability and feedback control at single synapses.
\newblock {\em Nature Reviews Neuroscience}, 10(5):373--383, 2009.

\bibitem{brunton2022data}
Steven~L Brunton and J~Nathan Kutz.
\newblock {\em Data-Driven Science and Engineering: Machine Learning, Dynamical Systems, and Control}.
\newblock Cambridge University Press, 2022.

\bibitem{cauwenberghs1992fast}
Gert Cauwenberghs.
\newblock A fast stochastic error-descent algorithm for supervised learning and optimization.
\newblock {\em Advances in Neural Information Processing Systems}, 5, 1992.

\bibitem{dalm2024efficient}
Sander Dalm, Joshua Offergeld, Nasir Ahmad, and Marcel van Gerven.
\newblock Efficient deep learning with decorrelated backpropagation.
\newblock {\em ArXiv preprint arXiv:2405.02385}, 2024.

\bibitem{davies2021advancing}
Mike Davies, Andreas Wild, Garrick Orchard, Yulia Sandamirskaya, Gabriel A~Fonseca Guerra, Prasad Joshi, Philipp Plank, and Sumedh~R Risbud.
\newblock Advancing neuromorphic computing with loihi: A survey of results and outlook.
\newblock {\em Proceedings of the IEEE}, 109(5):911--934, 2021.

\bibitem{dembo1990model}
A.~Dembo and T.~Kailath.
\newblock Model-free distributed learning.
\newblock {\em IEEE Transactions on Neural Networks}, 1(1):58--70, 1990.

\bibitem{Doob1942Brownian}
J~L Doob.
\newblock The {B}rownian movement and stochastic equations.
\newblock {\em Annals of Mathematics}, 43, 1942.

\bibitem{Faisal2008}
A~A Faisal, L~P~J Selen, and D~M Wolpert.
\newblock Noise in the nervous system.
\newblock {\em Nature Reviews. Neuroscience}, 9:292--303, 2008.

\bibitem{Feldbaum1960}
A~A Feldbaum.
\newblock Dual control theory, i.
\newblock {\em Avtomat. i Telemekh.}, 21:1240–1249, 1960.

\bibitem{fernando2009molecular}
Chrisantha~T Fernando, Anthony~ML Liekens, Lewis~EH Bingle, Christian Beck, Thorsten Lenser, Dov~J Stekel, and Jonathan~E Rowe.
\newblock Molecular circuits for associative learning in single-celled organisms.
\newblock {\em Journal of the Royal Society Interface}, 6(34):463--469, 2009.

\bibitem{Fernandez2024}
Jesús~García Fernández, Sander Keemink, and Marcel van Gerven.
\newblock Gradient-free training of recurrent neural networks using random perturbations.
\newblock {\em Frontiers in Neuroscience}, 18:1439155, 2024.

\bibitem{fiete2006gradient}
Ila~R Fiete and H~Sebastian Seung.
\newblock Gradient learning in spiking neural networks by dynamic perturbation of conductances.
\newblock {\em Physical Review Letters}, 97(4):048104, 2006.

\bibitem{flower1992summed}
Barry Flower and Marwan Jabri.
\newblock Summed weight neuron perturbation: An {O}(n) improvement over weight perturbation.
\newblock {\em Advances in Neural Information Processing Systems}, 5, 1992.

\bibitem{gagliano2016learning}
Monica Gagliano, Vladyslav~V Vyazovskiy, Alexander~A Borb{\'e}ly, Mavra Grimonprez, and Martial Depczynski.
\newblock Learning by association in plants.
\newblock {\em Scientific Reports}, 6(1):38427, 2016.

\bibitem{haider2021latent}
Paul Haider, Benjamin Ellenberger, Laura Kriener, Jakob Jordan, Walter Senn, and Mihai~A Petrovici.
\newblock Latent equilibrium: A unified learning theory for arbitrarily fast computation with arbitrarily slow neurons.
\newblock {\em Advances in Neural Information Processing Systems}, 34:17839--17851, 2021.

\bibitem{hiratani2022stability}
Naoki Hiratani, Yash Mehta, Timothy Lillicrap, and Peter~E Latham.
\newblock On the stability and scalability of node perturbation learning.
\newblock {\em Advances in Neural Information Processing Systems}, 35:31929--31941, 2022.

\bibitem{Jaeger2023}
Herbert Jaeger, Beatriz Noheda, and Wilfred~G. van~der Wiel.
\newblock Toward a formal theory for computing machines made out of whatever physics offers.
\newblock {\em Nature Communications 2023 14:1}, 14:1--12, 2023.

\bibitem{Kaspar2021}
C.~Kaspar, B.~J. Ravoo, W.~G. van~der Wiel, S.~V. Wegner, and W.~H.P. Pernice.
\newblock The rise of intelligent matter.
\newblock {\em Nature}, 594:345--355, 6 2021.

\bibitem{kidger2021on}
Patrick Kidger.
\newblock {\em {O}n {N}eural {D}ifferential {E}quations}.
\newblock PhD thesis, University of Oxford, 2021.

\bibitem{kirk1992analog}
David~B Kirk, Douglas Kerns, Kurt Fleischer, and Alan Barr.
\newblock Analog {VLSI} implementation of multi-dimensional gradient descent.
\newblock {\em Advances in Neural Information Processing Systems}, 5, 1992.

\bibitem{Kudithipudi2022}
Dhireesha Kudithipudi, Mario Aguilar-Simon, Jonathan Babb, Maxim Bazhenov, Douglas Blackiston, Josh Bongard, Andrew~P. Brna, Suraj~Chakravarthi Raja, Nick Cheney, Jeff Clune, Anurag Daram, Stefano Fusi, Peter Helfer, Leslie Kay, Nicholas Ketz, Zsolt Kira, Soheil Kolouri, Jeffrey~L. Krichmar, Sam Kriegman, Michael Levin, Sandeep Madireddy, Santosh Manicka, Ali Marjaninejad, Bruce McNaughton, Risto Miikkulainen, Zaneta Navratilova, Tej Pandit, Alice Parker, Praveen~K. Pilly, Sebastian Risi, Terrence~J. Sejnowski, Andrea Soltoggio, Nicholas Soures, Andreas~S. Tolias, Darío Urbina-Meléndez, Francisco~J. Valero-Cuevas, Gido~M. van~de Ven, Joshua~T. Vogelstein, Felix Wang, Ron Weiss, Angel Yanguas-Gil, Xinyun Zou, and Hava Siegelmann.
\newblock Biological underpinnings for lifelong learning machines.
\newblock {\em Nature Machine Intelligence 2022 4:3}, 4:196--210, 3 2022.

\bibitem{LeCun2015}
Yann Lecun, Yoshua Bengio, and Geoffrey Hinton.
\newblock Deep learning.
\newblock {\em Nature}, 521:436--444, 2015.

\bibitem{legenstein2010reward}
Robert Legenstein, Steven~M Chase, Andrew~B Schwartz, and Wolfgang Maass.
\newblock A reward-modulated hebbian learning rule can explain experimentally observed network reorganization in a brain control task.
\newblock {\em Journal of Neuroscience}, 30(25):8400--8410, 2010.

\bibitem{lillicrap2020backpropagation}
Timothy~P Lillicrap, Adam Santoro, Luke Marris, Colin~J Akerman, and Geoffrey Hinton.
\newblock Backpropagation and the brain.
\newblock {\em Nature Reviews Neuroscience}, 21(6):335--346, 2020.

\bibitem{Linnainmaa1970}
Seppo Linnainmaa.
\newblock {The representation of the cumulative rounding error of an algorithm as a Taylor expansion of the local rounding errors}.
\newblock {\em Master's Thesis (in Finnish), Univ. Helsinki}, pages 6--7, 1970.

\bibitem{lippe1994study}
D~Lippe and Joshua Alspector.
\newblock A study of parallel perturbative gradient descent.
\newblock {\em Advances in Neural Information Processing Systems}, 7, 1994.

\bibitem{lopez2023}
V\'ictor L\'opez-Pastor and Florian Marquardt.
\newblock Self-learning machines based on {H}amiltonian echo backpropagation.
\newblock {\em Physical Review X}, 13:031020, 8 2023.

\bibitem{Markovic2020}
Danijela Markovic, Alice Mizrahi, Damien Querlioz, and Julie Grollier.
\newblock Physics for neuromorphic computing.
\newblock {\em Nature Reviews Physics}, 2:499--510, 2020.

\bibitem{markram1997regulation}
Henry Markram, Joachim L{\"u}bke, Michael Frotscher, and Bert Sakmann.
\newblock Regulation of synaptic efficacy by coincidence of postsynaptic aps and epsps.
\newblock {\em Science}, 275(5297):213--215, 1997.

\bibitem{mcmahan2017communication}
H.~Brendan McMahan, Eider Moore, Daniel Ramage, Seth Hampson, and Blaise~Aguera y~Arcas.
\newblock Communication-efficient learning of deep networks from decentralized data.
\newblock In {\em Proceedings of the 20th International Conference on Artificial Intelligence and Statistics (AISTATS)}, volume~54, pages 1273--1282. PMLR, 2017.

\bibitem{Mead1990}
Carver Mead.
\newblock Neuromorphic electronic systems.
\newblock {\em Proceedings of the IEEE}, 78:1629--1636, 1990.

\bibitem{miconi2017biologically}
Thomas Miconi.
\newblock Biologically plausible learning in recurrent neural networks reproduces neural dynamics observed during cognitive tasks.
\newblock {\em Elife}, 6:e20899, 2017.

\bibitem{Modha2011}
Dharmendra~S Modha, Rajagopal Ananthanarayanan, Steven~K Esser, Anthony Ndirango, Anthony~J Sherbondy, and Raghavendra Singh.
\newblock Cognitive computing.
\newblock {\em Communications of the ACM}, 54:62--71, 2011.

\bibitem{Momeni2023}
Ali Momeni, Babak Rahmani, Matthieu Malléjac, Philipp del Hougne, and Romain Fleury.
\newblock Backpropagation-free training of deep physical neural networks.
\newblock {\em Science}, 11 2023.

\bibitem{money2021hyphal}
Nicholas~P Money.
\newblock Hyphal and mycelial consciousness: The concept of the fungal mind.
\newblock {\em Fungal Biology}, 125(4):257--259, 2021.

\bibitem{morrill2021neural}
James Morrill, Patrick Kidger, Lingyi Yang, and Terry Lyons.
\newblock Neural controlled differential equations for online prediction tasks.
\newblock {\em ArXiv preprint arXiv:2106.11028}, 2021.

\bibitem{Nakajima2022}
Mitsumasa Nakajima, Katsuma Inoue, Kenji Tanaka, Yasuo Kuniyoshi, Toshikazu Hashimoto, and Kohei Nakajima.
\newblock Physical deep learning with biologically inspired training method: gradient-free approach for physical hardware.
\newblock {\em Nature Communications}, 13:1--12, 12 2022.

\bibitem{oja1982simplified}
Erkki Oja.
\newblock Simplified neuron model as a principal component analyzer.
\newblock {\em Journal of Mathematical Biology}, 15:267--273, 1982.

\bibitem{payeur2021burst}
Alexandre Payeur, Jordan Guerguiev, Friedemann Zenke, Blake~A Richards, and Richard Naud.
\newblock Burst-dependent synaptic plasticity can coordinate learning in hierarchical circuits.
\newblock {\em Nature Neuroscience}, 24(7):1010--1019, 2021.

\bibitem{rao2001naive}
Venkatesh~G Rao and Dennis~S Bernstein.
\newblock Na{\"i}ve control of the double integrator.
\newblock {\em IEEE Control Systems Magazine}, 21(5):86--97, 2001.

\bibitem{rolls2010noisy}
Edmund~T Rolls and Gustavo Deco.
\newblock {\em The Noisy Brain: Stochastic Dynamics as a Principle of Brain Function}.
\newblock Oxford University Press, 2010.

\bibitem{rumelhart1985learning}
David~E Rumelhart, Geoffrey~E Hinton, and Ronald~J Williams.
\newblock Learning internal representations by error propagation.
\newblock Technical report, California Univ San Diego La Jolla Inst for Cognitive Science, 1985.

\bibitem{sasakura2013behavioral}
Hiroyuki Sasakura and Ikue Mori.
\newblock Behavioral plasticity, learning, and memory in c. {E}legans.
\newblock {\em Current Opinion in Neurobiology}, 23(1):92--99, 2013.

\bibitem{scellier2017equilibrium}
Benjamin Scellier and Yoshua Bengio.
\newblock Equilibrium propagation: Bridging the gap between energy-based models and backpropagation.
\newblock {\em Frontiers in Computational Neuroscience}, 11:24, 2017.

\bibitem{Schmidhuber1987}
Jürgen Schmidhuber.
\newblock {\em Evolutionary Principles in Self-Referential Learning}.
\newblock PhD thesis, 1987.

\bibitem{schultz1997neural}
Wolfram Schultz, Peter Dayan, and P~Read Montague.
\newblock A neural substrate of prediction and reward.
\newblock {\em Science}, 275(5306):1593--1599, 1997.

\bibitem{shahriari2015taking}
Bobak Shahriari, Kevin Swersky, Ziyu Wang, Ryan~P Adams, and Nando De~Freitas.
\newblock Taking the human out of the loop: A review of {B}ayesian optimization.
\newblock {\em Proceedings of the IEEE}, 104(1):148--175, 2015.

\bibitem{Shahsavari2023}
Mahyar Shahsavari, David Thomas, Marcel A.~J. van Gerven, Andrew Brown, and Wayne Luk.
\newblock Advancements in spiking neural network communication and synchronization techniques for event-driven neuromorphic systems.
\newblock {\em Array}, 20, 2023.

\bibitem{snoek2012practical}
Jasper Snoek, Hugo Larochelle, and Ryan~P Adams.
\newblock Practical {B}ayesian optimization of machine learning algorithms.
\newblock {\em Advances in Neural Information Processing Systems}, 25, 2012.

\bibitem{spall1992multivariate}
James~C Spall.
\newblock Multivariate stochastic approximation using a simultaneous perturbation gradient approximation.
\newblock {\em IEEE Transactions on Automatic Control}, 37(3):332--341, 1992.

\bibitem{Stern2021}
Menachem Stern, Daniel Hexner, Jason~W. Rocks, and Andrea~J. Liu.
\newblock Supervised learning in physical networks: From machine learning to learning machines.
\newblock {\em Physical Review X}, 11, 6 2021.

\bibitem{sutton2018reinforcement}
Richard~S Sutton.
\newblock {\em Reinforcement Learning: An Introduction}.
\newblock A Bradford Book, 2018.

\bibitem{Sutton2015}
Richard~S. Sutton and Andrew~G Barto.
\newblock {\em Reinforcement Learning: An Introduction}.
\newblock The MIT Press, 2nd edition, 2015.

\bibitem{tzen2019neural}
Belinda Tzen and Maxim Raginsky.
\newblock Neural stochastic differential equations: Deep latent {G}aussian models in the diffusion limit.
\newblock {\em ArXiv preprint arXiv:1905.09883}, 2019.

\bibitem{uhlenbeck1930theory}
George~E Uhlenbeck and Leonard~S Ornstein.
\newblock On the theory of the {B}rownian motion.
\newblock {\em Physical Review}, 36(5):823--841, 1930.

\bibitem{doremaele2024}
Eveline R.~W. Van~Doremaele, Tim Stevens, Stijn Ringeling, Simone Spolaor, Marco Fattori, and Yoeri van~de Burgt.
\newblock Hardware implementation of backpropagation using progressive gradient descent for in situ training of multilayer neural networks.
\newblock {\em Science Advances}, 10:8999, 7 2024.

\bibitem{wan2021learning}
Yi~Wan, Abhishek Naik, and Richard~S Sutton.
\newblock Learning and planning in average-reward markov decision processes.
\newblock In {\em International Conference on Machine Learning}, pages 10653--10662. PMLR, 2021.

\bibitem{Werbos1974}
P~Werbos.
\newblock {\em {Beyond Regression: New Tools for Prediction and Analysis in the Behavioral Sciences}}.
\newblock PhD thesis, 1974.

\bibitem{werfel2003learning}
Justin Werfel, Xiaohui Xie, and H~Seung.
\newblock Learning curves for stochastic gradient descent in linear feedforward networks.
\newblock {\em Advances in Neural Information Processing Systems}, 16, 2003.

\bibitem{whittington2017approximation}
James~CR Whittington and Rafal Bogacz.
\newblock An approximation of the error backpropagation algorithm in a predictive coding network with local {H}ebbian synaptic plasticity.
\newblock {\em Neural Computation}, 29(5):1229--1262, 2017.

\bibitem{whittington2019theories}
James~CR Whittington and Rafal Bogacz.
\newblock Theories of error back-propagation in the brain.
\newblock {\em Trends in Cognitive Sciences}, 23(3):235--250, 2019.

\bibitem{widrow199030}
Bernard Widrow and Michael~A Lehr.
\newblock 30 years of adaptive neural networks: perceptron, madaline, and backpropagation.
\newblock {\em Proceedings of the IEEE}, 78(9):1415--1442, 1990.

\bibitem{williams1992simple}
Ronald~J Williams.
\newblock Simple statistical gradient-following algorithms for connectionist reinforcement learning.
\newblock {\em Machine Learning}, 8:229--256, 1992.

\bibitem{Xu2022}
Winnie Xu, Ricky T~Q Chen, Xuechen Li, and David Duvenaud.
\newblock Infinitely deep {B}ayesian neural networks with stochastic differential equations.
\newblock {\em ArXiv:2102.06559v4 [stat.ML]}, 2022.

\bibitem{zuge2023weight}
Paul Z{\"u}ge, Christian Klos, and Raoul-Martin Memmesheimer.
\newblock Weight versus node perturbation learning in temporally extended tasks: Weight perturbation often performs similarly or better.
\newblock {\em Physical Review X}, 13(2):021006, 2023.

\end{thebibliography}


\end{document}